\documentclass[runningheads]{llncs}
\usepackage{graphicx}
\usepackage[colorlinks]{hyperref}
\usepackage{amsmath}
\usepackage{amssymb}
\usepackage{dsfont}
\usepackage{cite}
\usepackage{subcaption} 
\usepackage{multirow}
\usepackage{booktabs}
\usepackage{array}
\usepackage{xcolor}
\usepackage{makecell}
\usepackage{diagbox}
\newcommand{\name}{{\texttt{ConCAD}}}

\newcommand{\PreserveBackslash}[1]{\let\temp=\\#1\let\\=\temp}
\newcolumntype{C}[1]{>{\PreserveBackslash\centering}p{#1}}
\newcolumntype{R}[1]{>{\PreserveBackslash\raggedleft}p{#1}}
\newcolumntype{L}[1]{>{\PreserveBackslash\raggedright}p{#1}}

\makeatletter
\def\@seccntformat#1{\@ifundefined{#1@cntformat}%
   {\csname the#1\endcsname\quad}  
   {\csname #1@cntformat\endcsname}
}
\let\oldappendix\appendix 
\renewcommand\appendix{%
    \oldappendix
    \newcommand{\section@cntformat}{\appendixname~\thesection\quad}
}
\makeatother
\newenvironment{frcseries}{\fontfamily{frc}\selectfont}{}

\begin{document}
%

\title{ConCAD: Contrastive Learning-based Cross Attention for Sleep Apnea Detection}
%
%
\author{Guanjie Huang \and Fenglong Ma}
%
\authorrunning{}
%
\institute{College of Information Sciences and Technology, Pennsylvania State University, PA, 16802, USA. 
\email{\{gzh8, fenglong\}@psu.edu}}
%
\maketitle              
\begin{abstract}
With recent advancements in deep learning methods, automatically learning deep features from the original data is becoming an effective and widespread approach. However, the hand-crafted expert knowledge-based features are still insightful. These expert-curated features can increase the model's generalization and remind the model of some data characteristics, such as the time interval between two patterns. It is particularly advantageous in tasks with the clinically-relevant data,  where the data are usually limited and complex. To keep both implicit deep features and expert-curated explicit features together, an effective fusion strategy is becoming indispensable. In this work, we focus on a specific clinical application, i.e., sleep apnea detection. In this context, we propose a contrastive learning-based cross attention framework for sleep apnea detection (named {\name}). The cross attention mechanism can fuse the deep and expert features by automatically assigning attention weights based on their importance. Contrastive learning can learn better representations by keeping the instances of each class closer and pushing away instances from different classes in the embedding space concurrently. Furthermore, a new hybrid loss is designed to simultaneously conduct contrastive learning and classification by integrating a supervised contrastive loss with a cross-entropy loss. 
Our proposed framework can be easily integrated into standard deep learning models to utilize expert knowledge and contrastive learning to boost performance. As demonstrated on two public ECG dataset with sleep apnea annotation, {\name} significantly improves the detection performance and outperforms state-of-art benchmark methods.

\end{abstract}
\section{Introduction}
According to the National Institutes of Health of USA, 50 to 70 million people have chronic sleep disorders~\cite{altevogt2006sleep}, and the sleep disorders of sleep can increase the risk of many related diseases, such as hypertension and cardiovascular pathologies~\cite{young2002epidemiology}. Sleep apnea is one of the most common sleep disorders, which is an abnormal respiratory activity repeatedly occurring during sleep. 
The current primary method for diagnosing sleep apnea requires the patient to record the polysomnogram (PSG) in a clinic setup, which is very inconvenient and belated. 
Thus, how to automatically and effectively detect sleep apnea is a challenge, especially in the earlier stages.


Towards this end, automatic sleep apnea detection methods \cite{jezzini2015ecg,wang2019sleep,almutairi2021detection,urtnasan2018automated,dey2018obstructive} have been developed to simplify the diagnostic procedure, which including traditional machine learning methods and deep learning methods. 
Existing studies~\cite{wang2019sleep, wang2017time, supratak2017deepsleepnet} have shown that deep learning models perform better than the traditional machine learning ones, which require expert knowledge to manually extract features. However, these hand-crafted expert features are still valuable and insightful.
While researching the most appropriate hand-crafted features is time-consuming, there are a number of hand-crafted features that can be leveraged right away, as summarized by previous studies over centuries. 
In this context, we proposed a \textbf{cross-attention mechanism} to combine the deep features and the hand-crafted features to take advantage of both of them appropriately.

On the other hand, the regular deep learning methods usually train with the cross-entropy (CE) loss for a classification task. Since the CE loss only focuses on learning the necessary features to solve the classification task over known training data, it can be easily impaired by the mislabeled data~\cite{zhang2018generalized}, which further hinders the quality of the learned representations~\cite{ouali2020spatial}. 
To alleviate the problem, a common solution is to collect more data so that the model can learn more general features without excessive discrimination. However, this solution is particularly impractical in clinically relevant data, such as electrocardiography (ECG), where the data are always limited, and the labeling is prone to human errors.
As a remedy, we design a novel hybrid loss that integrates the cross-entropy loss with a \textbf{contrastive loss}. The contrastive loss helps to learn more general and robust features by clustering similar data and pushing apart dissimilar ones.

To sum up, in this work, we proposed a novel CONtrastive learning-based Cross Attention for sleep apnea Detection ({\name}) using single ECG data. To the best of our knowledge, our work is the first to successfully integrate contrastive learning for sleep apnea detection. Our major contributions of this paper are as follows:
\begin{itemize}
    \item We propose a cross attention mechanism to combine the deep features and expert knowledge-based features, which automatically fuses the features by emphasizing the important parts based on each other synergistically.
    \item We design a novel hybrid loss that encompasses both the cross-entropy (CE) loss and the supervised contrastive (SC) loss. The SC loss help to learn more general and robust by minimizing the ratio of intra-class to inter-class similarity while cross-entropy CE loss focus on discovering the useful features to solve the classification task.
    \item We demonstrate state-of-the-art classification performance on two public ECG datasets outperforming all benchmark methods.
    \item We show that our proposed framework of contrastive learning-based cross attention has better generalization ability, especially when the number of labeled training data is limited, comparing to a naive deep learning method without it.
    \item Both the cross attention mechanism and contrastive learning can be painlessly integrated into standard deep learning models.
\end{itemize}


\section{Related Work}
\label{Related Work}
In this section, we review the studies related to the proposed {\name} model, including the work on sleep apnea detection, cross attention mechanism for feature fusion, and contrastive learning.

\subsection{Sleep Apnea Detection}
The standard approach to diagnosis sleep apnea requires the patient to sleep overnight at a clinic setup and record the polysomnography (PSG) by various physiological sensors, and then the outputs of PSG are visually inspected by a clinical expert to give a diagnosis \cite{bloch1997polysomnography, kapur2017clinical, nikolaidis2019augmenting}. This process is always inconvenient and uncomfortable. Thus, some studies have begun to simplify the procedure of diagnosing sleep apnea by only using a single physiological data, such as ECG~\cite{almutairi2021detection,dey2018obstructive}, EEG~\cite{almuhammadi2015efficient}, and the respiration signal \cite{van2018automated}. Among these physiological data, ECG is a less intrusive option and also strongly related to sleep apnea. 

To this end, several studies~\cite{jezzini2015ecg,varon2015novel} manually extract hand-crafted features and feed them to classifiers (e.g., random forest, support vector machine) for sleep apnea detection. Recently, with the development of deep learning methods, some studies~\cite{wang2019sleep, almutairi2021detection} extract RR interval (RRI) and the R-peak envelope (RPE) and build deep learning model to automatically learn representation and detect sleep apnea. Furthermore, several studies~\cite{dey2018obstructive,urtnasan2018automated} develop deep learning models to directly learn features from the raw ECG data and detect sleep apnea in an end-to-end style.

\subsection{Attention-based Feature Fusion}
Another line of related work is feature fusion, which aims at combining different features to obtain a more effective representation. The frequently-used feature fusion techniques are concatenation~\cite{supratak2017deepsleepnet}, summation~\cite{feichtenhofer2016convolutional}, and multiplication~\cite{wu2018and}. However, these operations evenly combine all the features together without considering the importance of each feature. Some of the features gathered will help the model make the right decision, while others can lead to significant misjudgment ~\cite{lin2020using}.

Recently, the usage of attention learning mechanism has shown remarkable performance improvement for different tasks, such as natural language processing~\cite{wang2016attention}, image classification~\cite{wang2017residual}, and object tracking~\cite{chu2017online}. The attention mechanism highlights the effective discriminant parts of features while suppressing the redundant parts to a certain degree. To further take advantage of the features extracted from multi-modality inputs, a cross attention mechanism has been proposed to derive an attention mask from different inputs mutually. In \cite{mohla2020fusatnet}, the authors use one modality (LiDAR) to generate an attention mask that controls the spatial features of a different modality (HSI). In \cite{hou2019cross}, the authors derive cross attention maps for each pair of class features and query sample features to highlight specific regions and make the extracted features more discriminative.

\subsection{Contrastive Learning}
All of the deep learning methods mentioned above are trained by the cross-entropy loss. The cross-entropy loss is the most commonly-used one in the classification tasks, which calculates the difference between the actual probability distribution of the data and the predicted probability distribution of the model \cite{rumelhart1986learning}. As we previously introduced, the cross-entropy loss has some limitations. Thus, a supervised contrastive loss is added as an auxiliary regularization to alleviate problems in our proposed framework. 

The contrastive loss has recently been widely used in self-surprised learning \cite{he2020momentum, oord2018representation, banville2020uncovering}, which aims at clustering the similar data and pushing apart the dissimilar data. A supervised version of the contrastive loss is proposed by \cite{khosla2020supervised} to leverage the label information. Their proposed supervised contrastive learning contains two steps: First, the supervised contrastive loss is used to learn a representation to cluster the data from the same class and separate the data from different classes; Second, they froze the model and add a multi-layer perceptron (MLP) as a classifier on its top for the classification task. Recently, supervised contrasting learning has been used for different applications, such as image classification \cite{khosla2020supervised}, few-shot classification\cite{ouali2020spatial}, and semantic segmentation \cite{wang2021exploring}.

\section{Methodology}
\label{Proposed Approach}
\begin{figure}[h]
  \centering
  \includegraphics[width=1.0\linewidth]{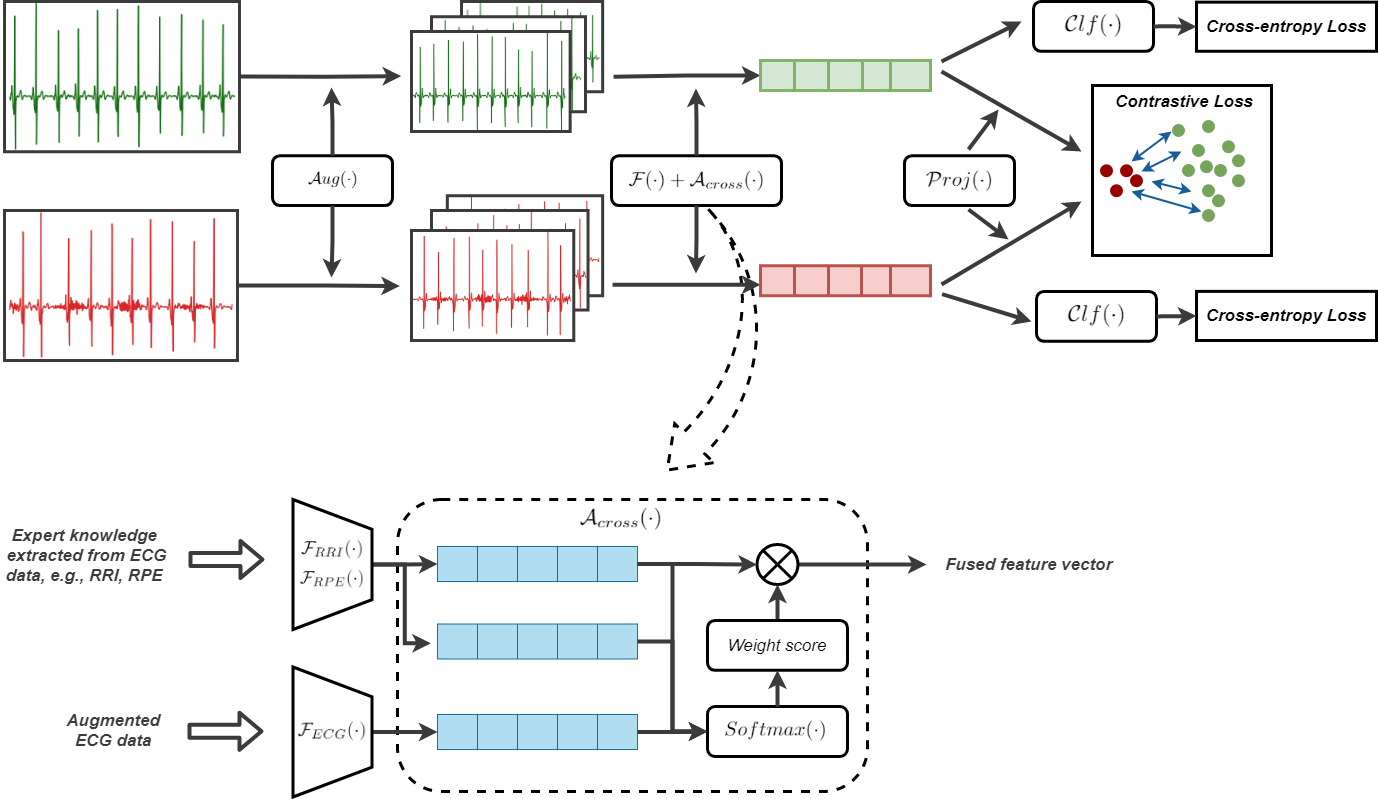}
  \caption{Overall of our proposed {\name} framework. The cross attention fuses the features by generating an attention weight mask. It can highlight the effective discriminant parts and suppress the irrelevant parts of features from ECG, RRI, and RPE collaboratively. Besides the cross-entropy, the supervised contrastive loss is also computed to optimize the intra-class to inter-class similarity ratio.}
  \label{fig:model_diagram}
\end{figure}

The goal of this work is to design an effective framework for leveraging the power of both the deep learning-based features and the expert knowledge-based features simultaneously to enhance the performance of sleep apnea detection. Towards this goal, we propose {\name} as shown in Fig. \ref{fig:model_diagram}, which is based on the contrastive learning framework to obtain better representations and utilizes a cross-attention mechanism to fuse different types of features.


Concretely, {\name} is achieved by three steps as shown in Fig. \ref{fig:model_diagram}. Firstly, the original raw data are passed through a feature extractor to learn the deep features. Simultaneously, the expert knowledge is passed through a feature extractor with a relatively shallow network to learn the expert features. Besides, data augmentation can be used before passing the data. Secondly, the deep features and the expert features are fed to a cross-attention module, which automatically fuses the features by emphasizing the important parts based on each other synergistically. Thirdly, the resultant attention-weighted features are mapped into a projection space for learning a representation with high intra-class similarity and low inter-class similarity to improve the classification accuracy by contrastive learning. Then, the learned representation is fed to the classification modules to output the probability of sleep apnea events.

As we are going to demonstrate the proposed {\name} on the datasets of ECG-based sleep apnea detection, the ECG data are certainly considered as the raw data input. Furthermore, the RRI and RPE manifest their effectiveness to detect sleep apnea by many research works \cite{al2009sleep, de2000automatic}. Consequently, the RRI and RPE are chosen as the expert knowledge input for our proposed framework.

\subsection{Expert Feature Extraction and Data Augmentation}
The expert knowledge is summarized by previous researches over the last centuries. In the field of detecting sleep apnea using ECG data, several previous studies \cite{al2009sleep, de2000automatic} have shown that the RR intervals (RRI) and R-peak envelope (RPE) are effective. To prepare the RRI and RPE data, we first detect the locations of the R-peaks by the Hamilton algorithm \cite{hamilton2002open}. Then, we calculate the distance between R-peaks as the RRI and use the amplitudes of the R-peaks as the RPE. Since the RRI can be easily disturbed by unexpected ECG spikes, a median filter is used to eliminate the disturbance as suggested by \cite{chen2014automatic}. Besides, since the number of the RRI or RPE is not always the same by giving a fixed time duration (e.g., 1 min), cubic interpolation was used to resample them to the same length \cite{wang2019sleep}. 

We are augmenting the ECG, RRI and RPE by two simple approaches: random time shift and reversion. Given the data $\textbf{x} = [x_0, x_1, x_2, \ldots,x_n]$, the random time shift will obtain  $\textbf{x}_{shift} = [x_t, x_{1+t}, x_{2+t}, \ldots,x_{n+t}]$, where $t$ is a randomly-generated number and represents the number of data points to shift. The revision will generate $\textbf{x}_{reverse} = [x_n, x_{n-1},  \ldots,x_2, x_1,x_0]$. The augmentation is conducted in each batch to provide more positives (i.e., instances with the same label) during batch training, which benefits a more robust clustering of the projection space. The augmentation process is presented as $\mathcal{A}{ug}(\textbf{x})$.

\subsection{Feature Extractor}
The feature extractor should be designed case by case. In this study, we have three feature extractors, which are used for learning features from the ECG, RRI and RPE separately, and named $\mathcal{F}_{ECG}$, $\mathcal{F}_{RRI}$ and $\mathcal{F}_{RPE}$. 

$\mathcal{F}$ consists of 4 convolution blocks. The first three blocks are made of one convolutional layer, one batch normalization layer, one ReLU activation layer, one maxpooling layer, and one dropout layer. The feature map size of the convolutional layer in the first block is chosen to cover data points of two contiguous beats in case that the patterns between beats get missed. The last convolution block does not contain the maxpooling and dropout layer. The module can be represented as $\textbf{x}'=\mathcal{F}(\textbf{x};\theta_{{F}})$, where $\textbf{x}$ represents the input data and $\theta_F$ denotes the parameters of the module.

Since there are three kinds of data, we have three corresponding extractors, which are
$\textbf{x}'_{ECG}=\mathcal{F}_{ECG}(\textbf{x}_{ECG};\theta_{{F}_{ECG}})$ for ECG data, $\textbf{x}'_{RRI}=\mathcal{F}_{RRI}(\textbf{x}_{RRI};\theta_{{F}_{RRI}})$ and  $\textbf{x}'_{RPE}=\mathcal{F}_{ECG}(\textbf{x}_{RPE};\theta_{{F}_{RPE}})$ for the expert knowledge-based features. More details of the extractors are described in Appendix \ref{app:a}.


\subsection{Cross Attention}
Not all the deep features and expert features contribute equally to the classification task. Thus, we design a cross-attention module, $\mathcal{A}_{cross}$, to collaboratively learn their importance and concentrate more on the important ones. The cross-attention is designed to ask the model to concentrate on the particular features, which contribute more to distinguish the instances from different classes. 
Before computing the cross attention, since the outputs of feature extractors are likely to have different dimensions, we need to project them to the same space by a linear transformation. Given $\textbf{x}'\in\mathbb{R}^{m\times{n}}$, the transformation is
\begin{equation}
    \textbf{x}''=\textbf{u}^\top\textbf{x}'\textbf{V} 
\end{equation}
where $\textbf{u}\in\mathbb{R}^{{m}}$ and $\textbf{V}\in\mathbb{R}^{n\times{k}}$ are trainable parameters. 

After it, $\textbf{x}''_{ECG}$, $\textbf{x}''_{RRI}$ and $\textbf{x}''_{RPE}$ have the same dimension $k$. Then, we are going to compute the attention weights. Specifically,
\begin{equation}
\begin{split}
    \boldsymbol{\alpha} &=Softmax([\alpha_{ECG},\alpha_{RRI},\alpha_{RPE}])\\
    \alpha_i &= \mathbf{w}_i^\top\mathbf{x}_i''+b_i
\end{split}
\end{equation}
where $i\in\mathcal{S} = \{ECG, RRI,RPE\}$ and $\boldsymbol{\alpha}\in\mathbb{R}^{3}$ is the attention weights. $\textbf{w}\in\mathbb{R}^{k}$ and $b\in\mathbb{R}$ are trainable parameters. The transformed $\textbf{x}''$ is passed through an one-layer MLP to learn the importance of different types of features synergistically, and the importance is normalized by a softmax function. 

Lastly, we compute the context vector \textbf{c} by
\begin{equation}
    \textbf{c} = \sum_{i\in\mathcal{S}}{\alpha_i\textbf{x}_i''}
\end{equation}
The context vector is the fused feature vector that is the weighted sum of features from different inputs based on the learned importance. The cross-attention module can be represented as $\mathcal{A}_{cross}([\textbf{x}'_{ECG}$, $\textbf{x}'_{RRI}$, $\textbf{x}'_{RPE}];\theta_{{A}_{cross}})$.

\subsection{Contrastive Learning}
For most of the conventional classification tasks, cross-entropy (CE) loss is commonly used to adjust model weights during training. However, CE loss may be impaired by noisy labels \cite{zhang2018generalized} and induce representations with excessive discrimination towards training data \cite{ouali2020spatial}. In order words, CE loss is likely to result in sub-optimal generalization.

As a remedy, contrastive learning is adopted to assist the model to learn more general and robust features by maximizing intra-class similarity while minimizing inter-class similarity. Concretely, we propose a novel hybrid loss, which utilizes the supervised contrastive (SC) loss \cite{khosla2020supervised} as an auxiliary regularization to the standard CE loss. 

\subsubsection{Contrastive Loss} 
The SC loss aims at simultaneously increasing the agreement among instances in positive pairs and encouraging the difference among instances in negative pairs. The instances with the same label form the positive pairs, and the instances with the different labels are considered as negative pairs. Specifically, the SC loss is computed in two steps.  We first project the input, i.e., the fused feature vector, to a lower dimension space by a one-layer MLP, and the low dimension vector is normalized to the unit hypersphere by L2 norm, $\textbf{z} = \mathcal{P}{roj}_{SC}(\mathcal{A}_{cross}([\textbf{x}'_{ECG}$, $\textbf{x}'_{RRI}$, $\textbf{x}'_{RPE}];\theta_{{A}_{cross}})$. Then, the SC loss can be computed by
\begin{equation}
    \mathcal{L}_{SC} = -\sum_{i=1}^{N}\frac{1}{N_{y_{i}}} \log \frac{\sum_{j=1}^{N} \mathds{1}_{[y_{i}=y_{j}]} \exp (\text{sim}(\textbf{z}_i, \textbf{z}_j) / \tau)}{\sum_{k=1}^{N} \mathds{1}_{[k \neq i]} \exp (\text{sim}(\textbf{z}_i, \textbf{z}_k) / \tau)},
\end{equation}
where $N$ is the batch size, and $N_{y_i}$ is the number of samples with the same label in each batch. $\mathds{1}_{[\cdot]}$ denotes an indicator function.  sim($\cdot$) represents the measure of similarity, and here the cosine similarity is used, i.e., $\text{sim}(u, v) = {u \cdot v}/{\lVert{u}\rVert \lVert{v}\rVert}$. $\tau$ is a hyperparameter that controls the strength of penalties on negative pairs \cite{wang2020understanding}.  

In the SC loss formula, the numerator represents the similarity of the positives, and the denominator represents the similarity of everything else in regard to $\textbf{z}_i$. The optimization of this formula pulls together the positives and pushes apart everything else.  That is, instances from the same class will form a closer cluster while the distances between clusters are increased in the projected hypersphere.   As a result, the model learns more general features instead of naively learning the features for the classification task over the known training data.

\subsubsection{Hybrid Loss}
As described in \cite{khosla2020supervised}, the standard SC loss requires two separate steps for a classification task: first, they train the feature extractor with the SC loss to learn a  representation vector; second, they freeze the feature extractor and train a classifier on the vector using the CE loss. However, the SC loss usually requires a very large batch size to achieve decent and stable performance. For example, \cite{khosla2020supervised} uses a batch size of 6,144. 
On the other hand, the CE loss only works in the second step and cannot update the model parameters in the feature extractor, which means that the CE loss does not make any contribution to learning the feature representation. 

To alleviate these problems, we use the SC loss as an auxiliary regularization term and integrate it with the CE loss. Specifically, we propose a new hybrid loss, which is the summation of CE and SC losses with a scaling parameter $\lambda$ to control the contribution of each loss:
\begin{equation}
    \mathcal{L}_{hybrid} = \lambda\mathcal{L}_{CE} + (1-\lambda)\mathcal{L}_{SC}.
\end{equation}

With the proposed hybrid loss, the model can take advantage of both the CE and SC losses simultaneously. The CE loss can learn effective features for classification tasks with small batch sizes, and the SC loss helps to promote these features to be more general and robust by minimizing the intra-class to inter-class similarity ratio. To train the model with the proposed hybrid loss, we project the fused feature vector to lower dimension hypersphere by $\mathcal{P}{roj}_{SC}(\cdot)$ to calculate the SC loss. At the same time,  the fused feature vector is sent to fully-connected layers $\mathcal{C}{lf}(\cdot)$ to calculate the CE loss as shown in the last step of Fig \ref{fig:model_diagram}.  All the layers except the last one in $\mathcal{C}{lf}(\cdot)$ use the ReLU activation, while the last layer is operated on the softmax activation, and its unit number needs to be identical to the number of classes. In addition, we will discard $\mathcal{P}{roj}_{SC}(\cdot)$ during prediction so that the proposed model has the same number of parameters as a model with only the CE loss.

\section{Experiments and Results}
\label{Experiments and Results}
\subsection{Datasets}
In our experiments, two datasets, i.e., Apnea-ECG \cite{penzel2000apnea} and  MIT-BIH Polysomnographic \cite{ichimaru1999development} obtained from Physionet \cite{goldberger2000physiobank}, are used for performance evaluation and comparison. Both datasets are publicly available and have been used to study sleep apnea detection methods in previous researches. 
\begin{itemize}
    \item \textbf{Apnea-ECG}: The apnea-ECG database is provided by Philipps University, which is the most commonly-used dataset for ECG-based sleep apnea studies. It contains 70 single-lead ECG recordings of varying lengths between 7 hours to 10 hours, sampled at the rate of 100 Hz. Each segment of 1 min ECG data is annotated by the expert as either apnea or normal event. The datasets are officially split into two sets by the provider: a released set of 35 recordings and a withheld set of 35 recordings. After removing the data with an unreasonable heartbeat rate,  the released set contains 16,888 segments and the withheld set contains 17,120 segments.
    \item \textbf{MIT-BIH Polysomnographic}: The MIT-BIH Polysomnographic database is collected by Boston's Beth Israel Hospital Sleep Laboratory. It contains over 80 hours of polysomnographic (PSG) recordings during sleep. Each recording includes a single channel of ECG annotated beat-by-beat and EEG and respiration signals. Each segment of 30 seconds of data is annotated with respect to sleep stages and apnea.  After removing the data with an unreasonable heartbeat rate, the final dataset contains 9,717 segments.
\end{itemize}

\subsection{Compared Methods}
To valid the performance of our framework, we use several state-of-the-art methods as our benchmark methods:
\begin{itemize}
    \item Support Vector Machine (\textbf{SVM}), Random Forest (\textbf{RF}), K-Nearest Neighbor (\textbf{KNN}), and Multi-Layer Perception (\textbf{MLP}) is adopted with 10 popular hand-crafted features from ECG data (e.g., RMSSD, NN50, etc.) as benchmark methods according to the work by \cite{jezzini2015ecg}.
    \item \textbf{LeNet-5}: In \cite{wang2019sleep}, a LeNet-5 convolutional neural network is used to learn features from RRI and RPE for sleep apnea detection. 
    \item \textbf{CNN+LSTM}: In \cite{almutairi2021detection}, three different deep learning architectures are proposed.  We adopt their best performing architecture, i.e., CNN+LSTM, as one of our benchmark methods.
    \item \textbf{ResNet}: In \cite{wang2017time},  a strong baseline model with the ResNet structure is proposed for time series classification, including ECG classification. So, we also use it for comparison. 
    \item \textbf{CNN-4}: In \cite{dey2018obstructive}, a four-layer CNN-based model with a novel pooling layer is proposed to detect sleep apnea from ECG data directly, and we compare it with our proposed method as well.
    \item \textbf{CNN-6}: In \cite{urtnasan2018automated}, several models with a different number of convolutional layers are designed to predict sleep apnea with ECG data. We adopt their best performing one, which contains 6 convolutional layers for our comparison.
\end{itemize}

We also compare the following approaches to show the improvements of the proposed framework step by step:
\begin{itemize}
    \item $\mathcal{F}_{ECG}+\mathcal{C}{lf}$: It employs a CNN-based feature extractor to learn the deep features from the raw ECG data and then classifies the targets by several fully-connected layers. It can be considered as a standard architecture of a naive deep learning model.
    \item $\mathcal{F}_{ECG} + \mathcal{F}_{RRI} + \mathcal{F}_{RPE} +\mathcal{C}{lf}$: Besides the raw ECG data, RRI and RPE are used as expert knowledge inputs. A simple concatenation is used to combine the features from ECG, RRI, and RPE. Then the concatenated features are sent for classification. 
    \item $\mathcal{F}_{ECG} + \mathcal{F}_{RRI} + \mathcal{F}_{RPE} +\mathcal{A}_{cross}+\mathcal{C}{lf}$: Instead of using the simple concatenation, a cross-attention mechanism is proposed to collaboratively fuse the features from ECG, RRI, and RPE.
    \item $\mathcal{F}_{ECG} + \mathcal{F}_{RRI} + \mathcal{F}_{RPE} +\mathcal{A}_{cross}+\mathcal{P}{roj}_{SC}+\mathcal{C}{lf}$: The proposed hybrid loss is used to update the model's parameter by adding an auxiliary projection during training to learn more general and useful features.
    \item {\name} ($\mathcal{A}{ug}+\mathcal{F}_{ECG} + \mathcal{F}_{RRI} + \mathcal{F}_{RPE} +\mathcal{A}_{cross}+\mathcal{P}{roj}_{SC}+\mathcal{C}{lf}$): Data augmentation is used with the architecture mentioned above to help learn more general and robust features to boost performance.
\end{itemize}

\subsection{Experiment Setup}
For the Apnea-ECG dataset, we train all the models, including the proposed {\name} method and benchmark methods, on the released dataset and test them on the withheld dataset. For the MIT-BIH PSG dataset, 10-fold cross-validation is used to examine the performance as there is no predefined training and test set. Moreover, since some existing studies \cite{wang2019sleep,supratak2017deepsleepnet, yadollahi2009acoustic} have shown that adjacent segment information helps analyze the sleep-related problems, the labeled segment with its surrounding $\pm 2$ segments of the ECG data is also included in our study. Thus, we will examine segments of 1 and 5 min on the Apnea-ECG dataset and test segments of 0.5 and 2.5 min on the MIT-BIH PSG dataset. In addition, some of the deep learning-based benchmark methods (i.e., \cite{wang2017time,dey2018obstructive,urtnasan2018automated}) are modified by increasing the pooling size and replacing flatten layer with GlobalAveragePooling for the input of 5 min and 2.5 min as they do no have a version to handle data with adjacent segments, and processing very long vector with their original structures exceeds our hardware memory limitation.

The proposed {\name} model is trained by the AMSGrad optimizer \cite{reddi2019convergence}, and all its parameters are initialized using HeNormal initializer \cite{he2015delving}. An initial learning rate of 0.005 is chosen and it decreases to 0.001 after certain epochs (e.g. 200 epochs). Moreover, the L2 regularization is added to the feature extractor $\mathcal{F}$ to prevent the model from overfitting into the noise or artifacts. 

\subsection{Results and Discussions}

We compare the performance of {\name} with other state-of-the-art benchmark methods, and the results are listed in Table \ref{tab:benchmark_compare}. Our proposed framework achieves an accuracy of $88.75\%$ with the 1 min segment input and  $91.22\%$ with the 5 min segment input on the Apnea-ECG dataset, and  $82.50\%$ with the 1 min segment input and  $83.47\%$ with the 5 min segment input on the MIT-BIH PSG dataset, which outperforms other benchmark methods. Besides, we can see deep learning methods can adaptively learn features from a different length of input while the machine learning methods with hand-crafted features are more sensitive to the change of the input length. With the adjacent segments, the deep learning model can learn more effective features for classification tasks.

\begin{table}[]
\centering
\caption{Accuracy of the proposed framework with other state-of-the-art methods on Apnea-ECG and MIT-BIH PSG datasets.}
\label{tab:benchmark_compare}
\begin{tabular}{L{3.5cm}lllC{1.1cm}cC{1.1cm}cC{1.1cm}cC{1.1cm}} 
\toprule
\multicolumn{2}{c}{\multirow{2}{*}{Methods}}             & \multicolumn{1}{c}{\multirow{2}{*}{Ref}} & \multirow{2}{*}{} & \multicolumn{3}{c}{Apnea-ECG}        & \multicolumn{1}{l}{} & \multicolumn{3}{c}{MIT-BIH PSG}     \\ 
\cline{5-7}\cline{9-11}
\multicolumn{2}{c}{}                                     & \multicolumn{1}{c}{}                     &                   & 1 min & \multicolumn{1}{l}{} & 5 min & \multicolumn{1}{l}{} & 0.5 min & \multicolumn{1}{l}{} & 2.5 min  \\ 
\midrule
\multirow{4}{*}{\makecell[l]{Feature Based \\Machine Learning (ML)}} & SVM             & \multirow{4}{*}{\cite{jezzini2015ecg}}          & \multirow{4}{*}{} &74.57      &                    &  67.52      &                     &   70.02    &                      &  70.30     \\
                                       & RF              &                                          &                 &   74.86      &                     &  72.30       &                      &   70.54   &                      &  68.15    \\
                                       & KNN             &                                          &                   &  71.81     &                      &  67.80     &                      & 69.51     &                      &     68.12  \\
                                       & MLP             &                                          &                   &   74.81    &                  &    70.59   &                      &   71.28   &                      &      71.20\\ 
\midrule
\multirow{4}{*}{Deep Learning (DL)}    & LeNet-5         & \cite{wang2019sleep}                            &               & 83.17          &                      &    87.25   &                      &    72.49   &                     &     78.82  \\
                                       & CNN+LSTM        & \cite{almutairi2021detection}                   &                   &   82.77    &                      &    86.12   &                      &  75.80   &                      &    80.79 \\
                                       & ResNet          & \cite{wang2017time}                             &                   &  83.57     &                      &    85.33   &                      &  77.29    &                      &    79.23   \\
                                       & CNN-4             & \cite{dey2018obstructive}                    &                   &  81.65     &                     &    84.42    &                      &   73.56    &                      &    76.92   \\ 
                                       & CNN-6      & \cite{urtnasan2018automated}                    &                   &  82.12     &                      &    84.37   &                      &  79.69    &                      &   82.25   \\ 
\midrule
Proposed method                              & {\name} &                                          &                   & \textbf{88.75} &                      & \textbf{91.22} &                      & \textbf{ 82.50}    &                      &     \textbf{83.47}  \\
\bottomrule
\end{tabular}
\end{table}

\begin{table}[h]
\centering
\caption{Accuracy of different architectures of the proposed framework on Apnea-ECG and MIT-BIH PSG datasets}
\label{tab:baseline_compare}
\begin{tabular}{llccccccc} 
\toprule
\multicolumn{1}{c}{\multirow{2}{*}{Architectures}} & \multirow{2}{*}{} & \multicolumn{3}{c}{Apnea-ECG}        & \multicolumn{1}{l}{} & \multicolumn{3}{c}{MIT-BIH PSG}     \\ 
\cline{3-5}\cline{7-9}
\multicolumn{1}{c}{}                         &                   & 1 min & \multicolumn{1}{l}{} & 5 min & \multicolumn{1}{l}{} & 0.5 min & \multicolumn{1}{l}{} & 2.5 min  \\ 
\midrule
$\mathcal{F}_{ECG}+\mathcal{C}{lf}$                                      & \multirow{4}{*}{} & 83.41 &                      &  85.48     &                      &  78.83     &                      &   80.11     \\
$\mathcal{F}_{ECG} + \mathcal{F}_{RRI} + \mathcal{F}_{RPE} +\mathcal{C}{lf}$                                      &                   & 83.23 &                      &  87.64     &                      &   79.42   &                      &     80.60 \\
$\mathcal{F}_{ECG} + \mathcal{F}_{RRI} + \mathcal{F}_{RPE} +\mathcal{A}_{cross}+\mathcal{C}{lf}$                                     &                   & 85.35 &                      & 89.43      &                      &   80.22  &                      &     81.77  \\
$\mathcal{F}_{ECG} + \mathcal{F}_{RRI} + \mathcal{F}_{RPE} +\mathcal{A}_{cross}+\mathcal{P}{roj}_{SC}+\mathcal{C}{lf}$                                      &                   & 87.16 &                      &   90.85    &                      &    81.83  &                      &    82.83   \\
{\name}&                   & \textbf{88.75} &                      & \textbf{91.22} &                      &   \textbf{82.50}   &                      &     \textbf{83.47}  \\
\bottomrule
\end{tabular}
\end{table}

We also examine the proposed framework step by step to show the effectiveness of each step in Table \ref{tab:baseline_compare}. We can see that the performance can be worse if we simply concatenate the deep features with the expert features as some of the features will help the model make the better judgment possible, while others are likely to act as noise and thereby lead to more errors. With the cross attention module $\mathcal{A}_{cross}$, we can see that the model learns a better-fused feature representation by learning an attention mask synergistically from each other. The new fused feature representation maintains the effective discriminant parts of features while suppressing the irrelevant parts. Specifically, the accuracy improves from $83.41\%$ to $85.35\%$ with the 1 min segment input and from $85.48\%$ to $89.43\%$ with the 5 min segment input on the Apnea-ECG dataset. On the MIT-BIH PSG dataset, the accuracy improves from $78.83\%$ to $80.22\%$ with the 0.5 min segment input and from $80.11\%$ to $81.77\%$ with the 2.5 min segment input.

In Table \ref{tab:baseline_compare}, we can also see a further improvement by using the proposed hybrid loss, which takes advantage of both CE loss and SC loss. The accuracy increases to $87.16\%$ with the 1 min segment input and  $89.43\%$ with the 5 min segment input on the Apnea-ECG dataset. On the MIT-BIH PSG dataset, the accuracy increases to $81.83\%$ with the 1 min segment input and  $81.77\%$ with the 5 min segment input. The hybrid loss boosts the performance by promoting the model to learn more general and discriminant feature representations in case that the model overfits into the training data by learning features with excessive discrimination. 

\begin{figure}[h]
    \centering
    \begin{subfigure}[t]{0.48\textwidth}
        \centering
        \includegraphics[width=\textwidth]{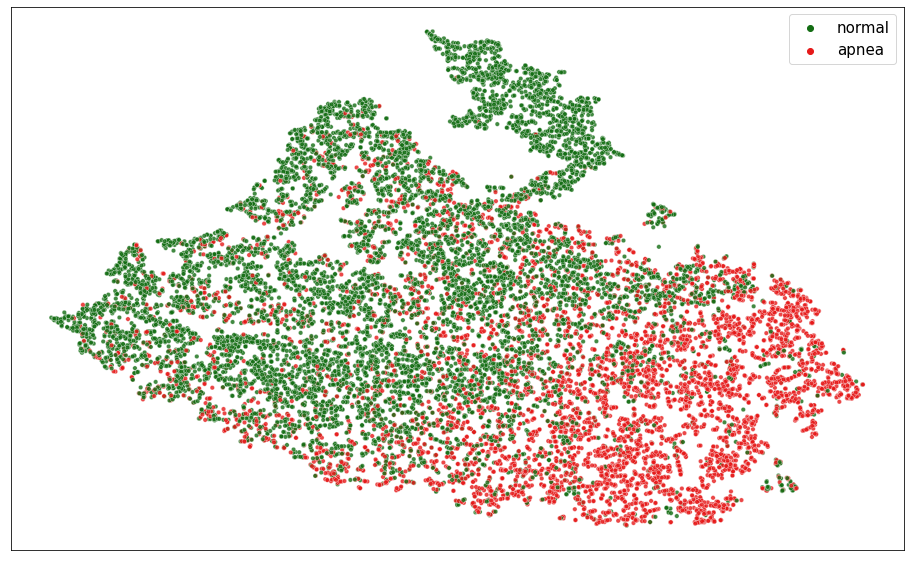}
        \caption{CE loss}
        \label{fig:apnea_ecg_no_contrastive}
    \end{subfigure}
    \smallskip
    \begin{subfigure}[t]{0.48\textwidth}
        \centering
        \includegraphics[width=\textwidth]{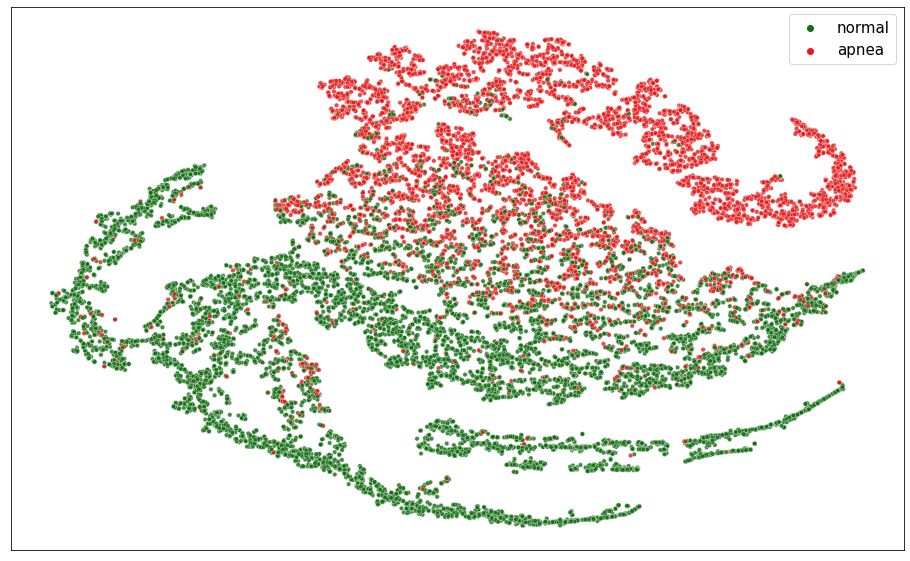}
        \caption{Proposed hybrid loss}
        \label{fig:apnea_ecg_with_contrastive}
    \end{subfigure}
    \\
    \begin{subfigure}[t]{0.48\textwidth}
        \centering
        \includegraphics[width=\textwidth]{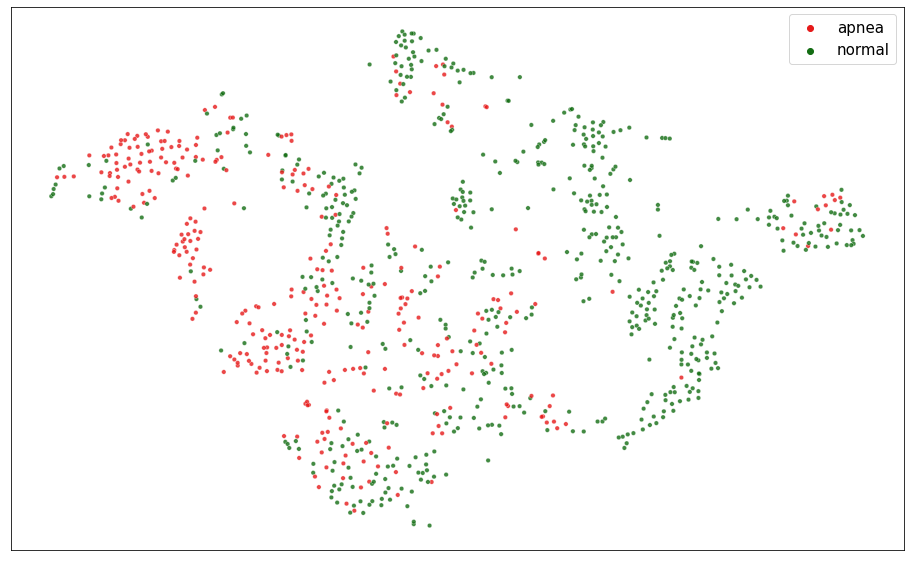}
        \caption{CE loss}
        \label{fig:mitbih_no_contrastive}
    \end{subfigure}
    \smallskip
    \begin{subfigure}[t]{0.48\textwidth}
        \centering
        \includegraphics[width=\textwidth]{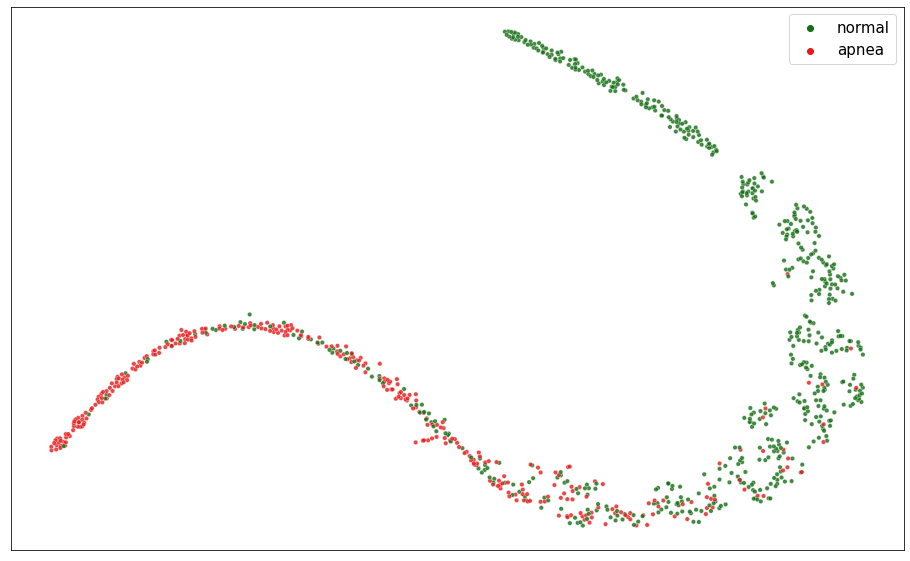}
        \caption{Proposed hybrid loss}
        \label{fig:mitbih_with_contrastive}
    \end{subfigure}
    
    \caption{The t-SNE plots of the fused feature vector on the withheld set of Apnea-ECG (\subref{fig:apnea_ecg_no_contrastive} and \subref{fig:apnea_ecg_with_contrastive}) and the validation set of MIT-BIH PSG (\subref{fig:mitbih_no_contrastive} and \subref{fig:mitbih_with_contrastive}), comparing the cross-entropy (CE) loss with the proposed hybrid loss.}
    \label{fig:tsne}
\end{figure}

In Fig. \ref{fig:tsne}, the t-SNE plots show the learned feature representation with the CE loss and the proposed hybrid loss. We can see that the hybrid loss promotes to more compact clustering of the instances from the same class while the representation with CE loss is more scattered.
We also think the attention module benefits contrastive learning by focusing on parts of features when increasing the agreement among instances in positive pairs and encouraging the difference among instances in negative pairs. It is similar to human behavior that human usually tends to recognize an unseen data by comparing the most relevant parts with known ones. By using data augmentation, the proposed model can learn more general feature representation with a more clear boundary and achieve better performance.

\begin{figure}[h]
    \centering
    \begin{subfigure}[t]{0.48\textwidth}
        \centering
        \includegraphics[width=\textwidth]{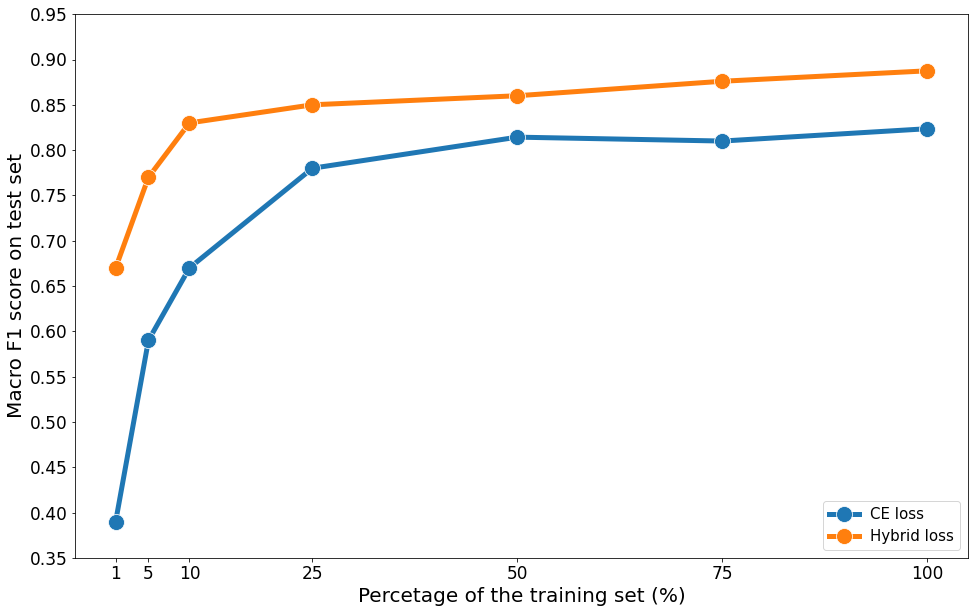}
        \caption{Apnea-ECG dataset}
        \label{fig:apnea_ecg_less_sample}
    \end{subfigure}
    \smallskip
    \begin{subfigure}[t]{0.48\textwidth}
        \centering
        \includegraphics[width=\textwidth]{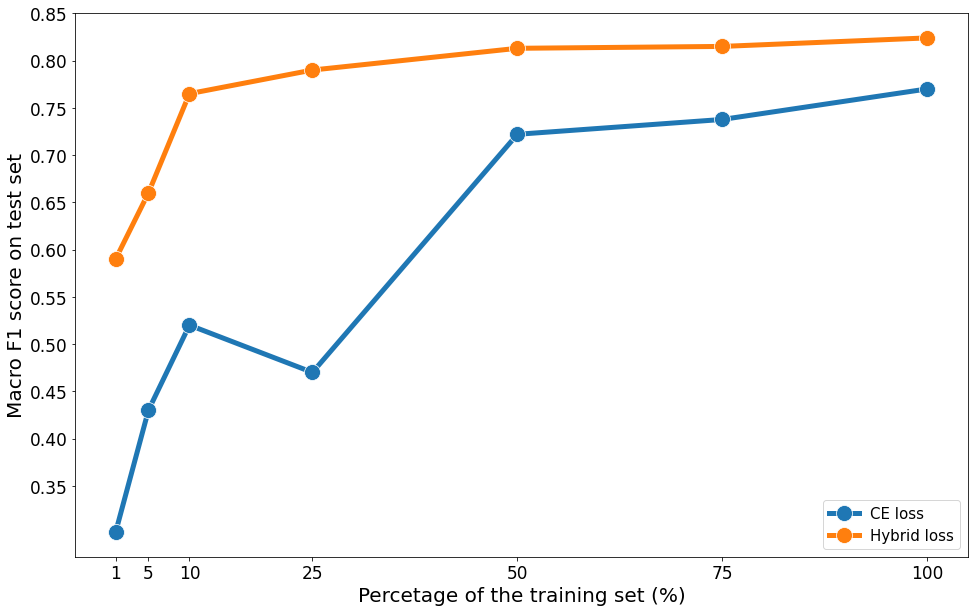}
        \caption{MIT-BIH PSG dataset}
        \label{fig:mitbih_psg_less_sample}
    \end{subfigure}
    
    \caption{Impact of number of training data on the performance of the sleep apnea detection with the cross entropy (CE) loss and the proposed hybrid loss. }
    \label{fig:less_sample}
\end{figure}

In addition, the hybrid loss enables the classification tasks with limited training labeled data. The limitation of labeled data is a prevalent and critical problem in the healthcare field. We train the model by using a fraction of the training set and test it on the entire test set.
The results are shown in Fig. \ref{fig:less_sample} in terms of the macro F1 score. F1 score can clearly show the quality of the model when the dataset is imbalanced. With only $1\%$ of the training data, the proposed {\name} model can still achieve an F1 score of  $0.67$ on the Apnea-ECG dataset and $0.59$ on the MIT-BIH PSG dataset. However, the CE loss performs poorly and skews into the majority class. Furthermore, the proposed model only requires $10\%$ of the training data to make a reasonable classification while the naive deep learning model needs more than $50\%$ to get decent performance. Hence, the proposed model with hybrid loss largely outperforms a naive deep learning approach with CE loss on smaller datasets.

\section{Conclusions and Future Work}
\label{Conclusions}
In this paper, we propose a contrastive learning-based cross attention framework, named {\name}. The cross attention leverages the expert knowledge and fuses it with deep features by highlighting each other collaboratively. The novel hybrid loss that encompasses the cross-entropy loss and supervised contrastive loss helps learn more robust features by clustering the same class data and pushing apart data of different classes in projection space. Moreover, we show the proposed framework achieves state-of-the-art results on two public ECG datasets. Furthermore, we show that the proposed framework has better generalization ability with limited labeled training data. We conclude that the ECG data with adjacent segments helps to detect the sleep apnea occurrence through the experiment.

In future work, we plan to study more ECG data augmentation techniques that would help contrastive learning to generate better representations. We also plan to develop more interfaces to allow different formats of expert knowledge (e.g., electronic health records) to be integrated into our framework.

%
%
\bibliographystyle{splncs04}
\bibliography{ref}

\appendix
\section{} 
\label{app:a}

The feature extractor are different for different data and tasks. In this study, we design a CNN-based extractors for ECG, RRI and RPE separately. The structure of the extractor for two dataset are also different as their ECG data have different sampling frequency and noise. The details are shown in the table below. The ConvBlock(number of filters, kernel size, stride) is made of one convolutional layer, one batch normalization layers, one ReLU activation layer.

\begin{table}
\centering
\small
\caption{The details of the feature extractors used for ECG, RRI and RPE on Apnea-ECG and MIT-BIH PSG.}
\begin{tabular}{L{3.8cm}|L{4.9cm}|L{3.8cm}} 
\toprule
$\mathcal{F}_{ECG}$ (Apnea-ECG)                                                                                                                                   & $\mathcal{F}_{ECG}$ (MIT-BIH PSG)                                                                                                                                                                                           & $\mathcal{F}_{RRI}$, $\mathcal{F}_{RPE}   $                                                                                                                                            \\ 
\midrule
ConvBlock(64,100,20)-MaxPool(2)-Dropout(0.5)-ConvBlock(64,8,4)-MaxPool(2)-Dropout(0.5)-ConvBlock(128,4,2)-MaxPool(2)-Dropout(0.5)-ConvBlock(128,4,2) & ConvBlock(64,60,5)-MaxPool(2)-Dropout(0.5)-ConvBlock(128,8,3)-ConvBlock(128,8,3)-MaxPool(2)-Dropout(0.5)-ConvBlock(256,4,2)-ConvBlock(256,4,2)--MaxPool(2)-Dropout(0.5)-ConvBlock(128,4,1)-ConvBlock(128,4,1) & ConvBlock(64,8,4)-MaxPool(2)-Dropout(0.5)-ConvBlock(64,4,2)-MaxPool(2)-Dropout(0.5)-ConvBlock(128,2,1)-MaxPool(2)-Dropout(0.5)-ConvBlock(128,2,1)  \\
\bottomrule
\end{tabular}
\end{table}
\end{document}